\documentclass[runningheads]{llncs}
\usepackage{todonotes}
\usepackage{graphicx}
\usepackage[T1]{fontenc} 
\usepackage[utf8]{inputenc}
\usepackage{multirow}
\begin{document}
\title{Human-Centered Programming: The Design \\ of a Robotic Process Automation Language}
\titlerunning{The Design of a Human-Readable RPA Language}

% If the paper title is too long for the running head, you can set
% an abbreviated paper title here
%
\author{Piotr~Gago\inst{1}\orcidID{1111-2222-3333-4444} \and
Anna~Voitenkova\inst{1}\orcidID{1111-2222-3333-4444} \and
Daniel~Jablonski\inst{1}\orcidID{0000-1111-2222-3333} \and
Ihor~Debelyi\inst{1}\orcidID{1111-2222-3333-4444} \and
Kinga~Skorupska\inst{1}\orcidID{0000-0002-9005-0348} \and
Maciej~Grzeszczuk\inst{1}\orcidID{0000-0002-9840-3398} \and
Wieslaw~Kopec\inst{1}\orcidID{0000-0001-9132-4171}} 

\authorrunning{Gago et al.}
% First names are abbreviated in the running head.
% If there are more than two authors, 'et al.' is used.
%
\institute{Polish-Japanese Academy of Information Technology}

\maketitle              % typeset the header of the contribution
\begin{abstract}
RPA (Robotic Process Automation) helps automate repetitive tasks performed by users, often across different software solutions. Regardless of the RPA tool chosen, the key problem in automation is analyzing the steps of these tasks. This is usually done by an analyst with the possible participation of the person responsible for the given activity. However, currently there exists no one-size-fits-all description language, which would allow to record, process, and easily automate steps of specific tasks. Every RPA solution uses a different notation, which is not easily human-readable, editable, and which cannot be applied to a different automation platform. Therefore, in this paper, we propose a new eXtensible Robotic Language (XRL) that can be understood by both programmers and non-programmers to automate repetitive business processes.

\end{abstract}

\section{Introduction}

In our research, we explore the opportunities related to the development of new types of human-readable RPA tools. We follow the practices of people-oriented design to take into account the design aspects related to employee well-being \cite{kopec2018hybrid}. \textbf{Therefore, we aim to create a new eXtensible Robotic Language (XRL) that is human-readable and has a relatively low learning curve -- to empower people to engage with the co-creation of software robots and help them shape the technology landscape in the context of the growing impact of industry 4.0 and human-centered programming, including intelligent automation.} In this way, when business processes change, employees could anticipate these changes and make necessary adjustments to the RPA code. 
We hope that our proposal of a language, which is the result of collaboration between HCI experts and programmers \cite{chasins2021pl} may be easily understood by any English-speaking person, which will significantly facilitate the collection and analysis of data related to business processes.

\section{Related Work}

\subsection{Existing RPA Languages}

The current RPA tools, despite a very dynamic development, are still largely rule-based and recreate previously configured activities \cite{willcocks_paper_nodate} that are described and precisely defined. RPA tools such as UI Path\footnote{Automation platform developed by UiPath based on screen recording. Product page: https://www.uipath.com/}, Prism\footnote{RPA Automation software developed by Blue Prism Group based on drag-and-drop building. Product page: https://www.blueprism.com/}, or Selenium-based tools\footnote{A suite of automation tools. Product page: https://www.selenium.dev/} allow employees to automate business processes to some extent. \textbf{Yet, each RPA tool uses a different language to describe the business process.}  Each of these languages is specific to a particular technical RPA solution. One main problem we noticed about the output analysis from the leading RPA tools was that it was not human-readable. Therefore, the employees had to rely on existing GUI for the paid tools to make any changes, which may not be easy to implement, as to do that, they had to analyse the whole business process and find the relevant part to edit. The complexity of the  notation prevents users from conveniently editing or analyzing the saved process without dedicated tools. For this reason companies are tied to a particular tool. If they changed the tool, for example, to a cheaper one, they would have to pay re-implementation costs.\textbf{ If the RPA tools existing on the market had a common way of describing process automation, companies could migrate between tools without incurring major costs.} It should be noted that RPA tools are only used to automate business processes, not to optimize them. If a process contains redundant and suboptimal steps, the RPA tool will also perform those. \textbf{Task optimization could be more accessible if a consistent language could describe automated processes. This would also allow more freedom when changing or maybe even combining different tools.}

\subsection{Human-readable Language Heuristics}

We wanted to start with a definition of a concise and human-readable language that would allow us to describe the activities performed as part of the business process we want to automate. The key aspects of this language are:\\
\begin{enumerate}
\item Expressiveness. We should be able to express the activities performed as part of the automated process in a detailed way.
\item Extensibility. The language grammar cannot be closed and should take into account further changes as needed.
\item Computer processability. We must be able to process the proposed language automatically. This element is necessary if we assume that the language can later become the basis for developing a new class of tools for recording, reproducing, and analyzing automated business processes.
\item Human readability. We want to lower the barrier of entry to edit the code without additional tools.
\end{enumerate}

Although the first criteria are the result of formal requirements for this type of language, human-readability is something we wanted to verify in our present study, based on related works \cite{codereadability,codereadability2020,codereadabilitytesting2016,criteriaforevaluation2016}.

\section{Methods: How We Developed the Language} \label{seclangdev}

\subsection{Industry Collaboration: Business Processes}

The language development process was based on three different repetitive business processes performed by company employees. The scenarios were prepared in cooperation with a company that is professionally involved in automating business processes. Following established market practice, the employees' screens were recorded to show the steps of these processes, and the resulting videos served as a starting point for our work.

\begin{enumerate}
\item The first process combined the interaction between two applications: a web browser and a desktop application (Total Commander). The goal was to find a specific phrase using Total Commander, copy it to the clipboard, and send it to a recipient via Gmail in the browser.
\item The second was based on SQL Server Management Studio. Its goal was to generate a PDF report on disk usage for a selected database.
\item The last process focused on retrieving data from an Excel file stored locally and using these data in a CRM system (CRM Vision\footnote{Product page: https://crmvision.pl/}) later. The goal was to insert the data from the file to the CRM into appropriate fields in a form.
\end{enumerate}

\subsection{Industry Collaboration: Existing Programs}

The programs mentioned previously were chosen based on the existing problems and requirements of the market.
\begin{enumerate}
    \item "Total Commander" is a program originally coded using Delphi. The problem with such solutions is that it is difficult to access elements of the GUI using selectors and identifiers.
    \item "Microsoft SQL Management Studio" has UI developed in Windows Presentation Foundation (WPF). Newer technologies, like WPF, are simpler to work with, offering convenient access via API. On the other hand, there is a problem with accessing elements in different windows that appear when using the application.
    \item CRM Vision was chosen because it is a common scenario to automate particular tasks in CRM systems. If the CRM is written using web technologies, a browser is needed to use it. Therefore, the last business case covered both automating CRM systems and automating any web page or web application in the browser.
\end{enumerate}

\subsection{Charting the Business Processes}

The goal of doing all 3 processes was to understand what operations are commonly used during process execution, and based on this knowledge to start the design of our language from the bottom-up. The next step was to create console applications that repeat the flow of selected business processes. The applications were written using C\# programming language and ready-to-use automation tools: Selenium for web browsers and Appium (using the Windows driver) for desktop applications. The created console applications allowed us to work on a relatively low level with each program used in the processes (Total Commander, SQL Management Studio, and the web browser of choice). Those helped to understand the strengths and limitations of the programs' architecture.

The next step was to represent the three selected business processes in the form of flow charts to visualize and analyze all the steps one by one. When the flow chart diagrams were created, we compared them with each other. Our goal was to identify common basic steps for each diagram and create elements for our own language based on them. When an examined element of the flowchart was too complex (did more than one action), we split this element into several, more generic steps - which more easily could be reused later. In case we were not able to create one generic step matching all types of applications, we saved it as a separate step or created a hierarchy of elements for a certain flowchart step. After this, existing flowcharts were drawn one more time using our own elements to check if it is possible to represent a process using actions of the newly built language. Such an approach gives the user the possibility to create a more complex task using basic steps of our language and save it as a step for further reuse in other places.

\begin{figure}[h]
  \centering
  \includegraphics[width=\linewidth]{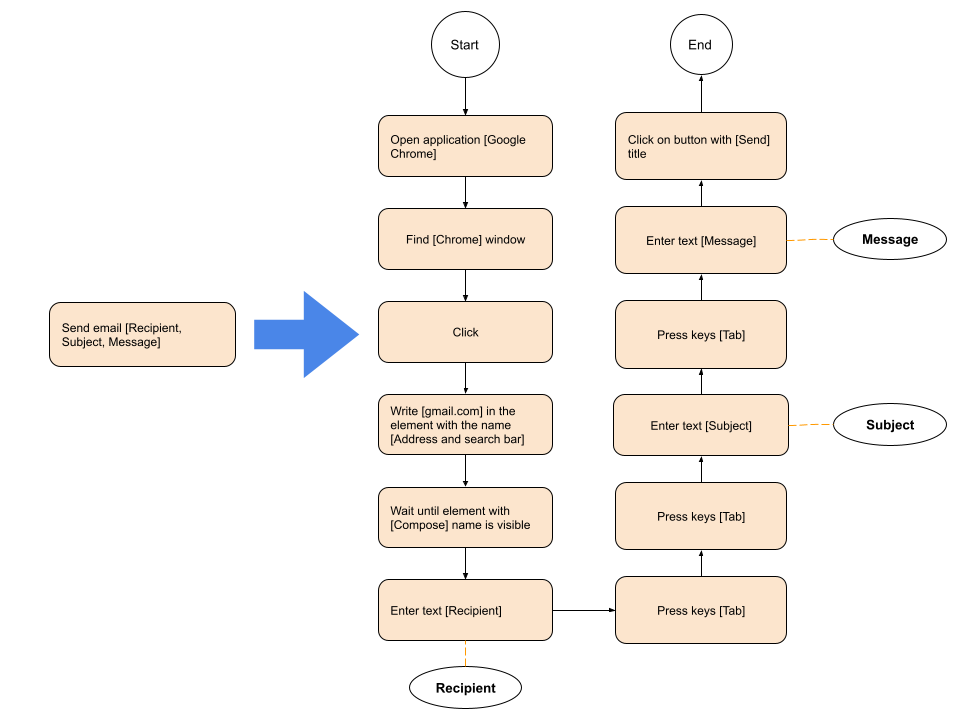}
  \caption{User action subdivided into essential steps}
\end{figure}

\section{Results: eXtensible Robotic Language (XRL): Our Proposed RPA Language and Representation Syntax}

    % \todo[inline,color=blue!55]{Review:
    % 1. Unfortunately, what missing in the paper is the detailed description of the
    % proposed language. This is especially problematic when making a comparison with
    % the two controlled XML-based solutions. It is understandable that the commercial
    % solutions should be anonymized. However, it would be much helpful to clarify the
    % major similarities and differences among these three solutions, in particular
    % those related to HCI.}

% \subsection{RPA language syntax}

Based on real business processes, we started the development of the basic syntax of our language. Choosing the primary format of the file was one of the first decisions to make. One of the options was to create a completely custom format. This would require implementing a custom parser and other tools later. Another option was to use formats such as XML, JSON, or YAML. We decided that using an already existing format and building a language from it will facilitate the subsequent processing of the language. Most programming languages have parsers and serializers that can work with these formats. Most current automation solutions often use XML-based languages. This format is theoretically human-readable. Unfortunately, the research carried out later confirmed that even with moderately easy processes, it becomes unreadable. We chose to use a format that offers a slightly tighter syntax and, therefore, much cleaner. For this purpose, we use the YAML (YAML Ain't Markup Language) format \cite{2021YamlOfficialPage} \footnote{YAML was created in 2001 by Clark Evans. It was a time of heavy use of declarative languages and problems related to them were increasingly noticed. YAML represents a data-oriented language, not a document-oriented language, which distinguishes it from XML-based solutions. It maintains simple syntax and readability even for people without previous knowledge.}. 

\subsection{Syntax examples}

\begin{figure}[h]
  \centering
  \includegraphics[width=0.8\linewidth]{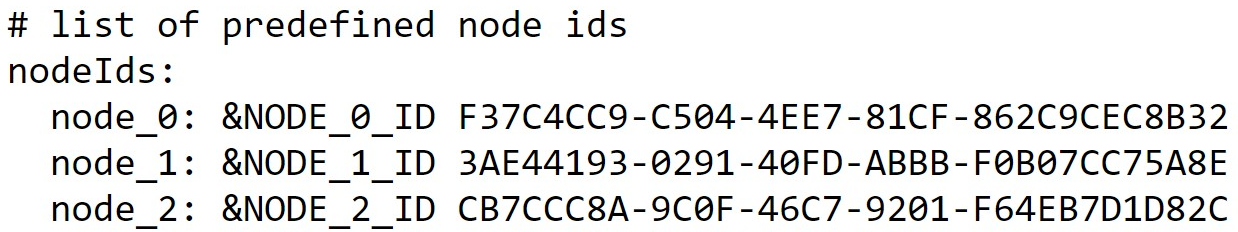}
  \caption{Fragment of the section used to define a list of identifiers}
  \label{firstsection}
\end{figure}

We started by defining a list of identifiers for individual actions in the process and called them nodes. An action may represent clicking on a button or entering text inside an element. All identifiers are placed in an array. Each ID has a name and an anchor (see Table \ref{firstsection}). An anchor allows us to refer to a given value elsewhere in our document, which reduces the need to copy the values of the identifiers. As identifiers, we used values compatible with the GUID (Globally Unique Identifier) format. This allows us to reduce duplication in the record of our process. We can refer to the value determined with the help of the anchor by the sign *.

\begin{figure}[htb!]
  \centering
  \includegraphics[width=0.8\linewidth]{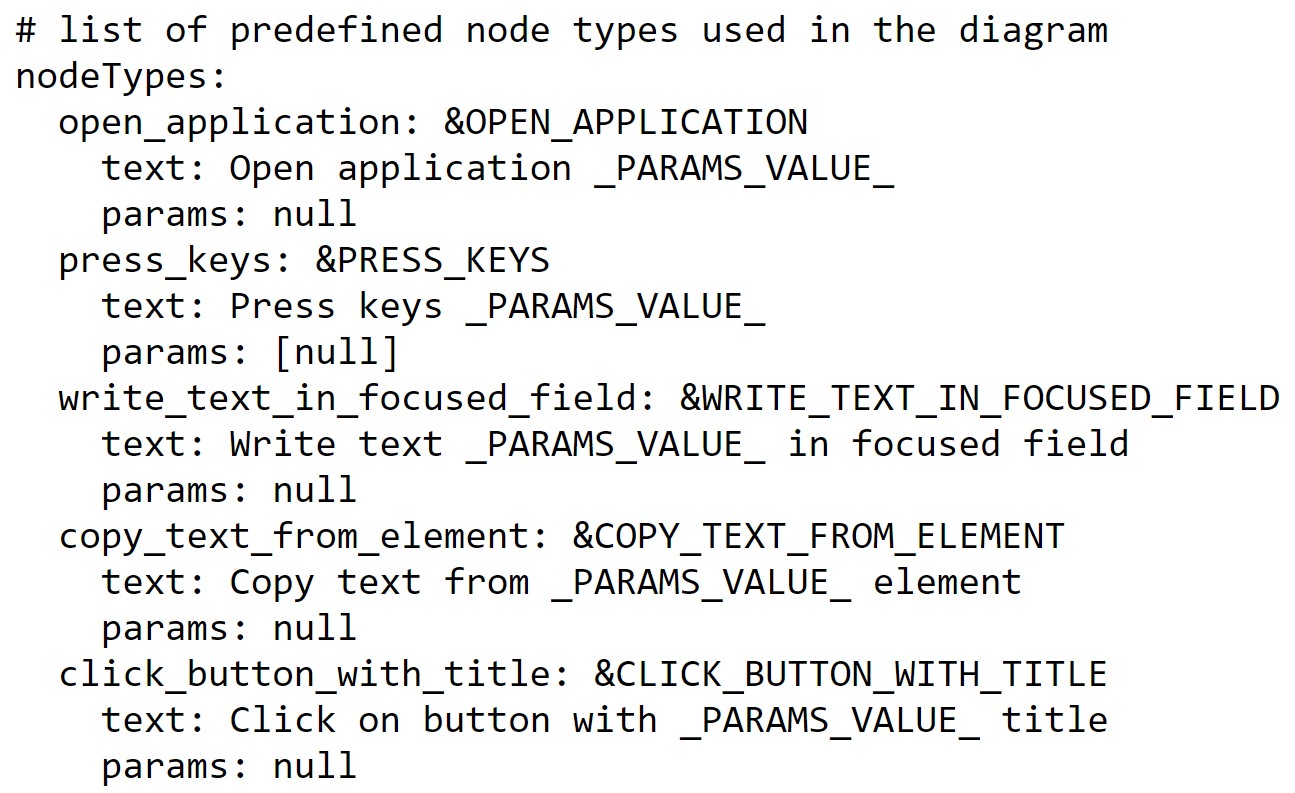}
  \caption{Section used to define a list of element types}
  \label{secondsection}
\end{figure}

\begin{figure}[h!]
  \centering
  \includegraphics[width=0.5\linewidth]{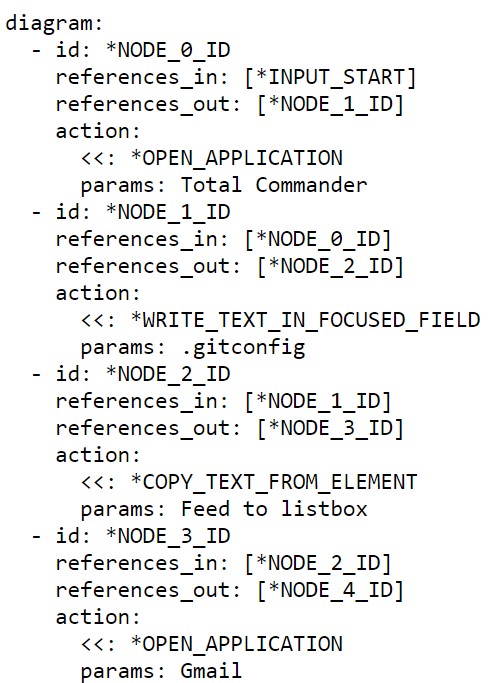}
  \caption{Section used to define the process using elements defined in previous sections}
  \label{fourthsection}
\end{figure}

Next, we define a list of element types that we want to use in a given process as an array (see Table \ref{secondsection}). We want to introduce a minimum set of elements as the basis for a given language. With their help, users could create more complex elements that are part of their processes. For example, with the help of simple actions such as clicking and entering text, we can build more complex elements such as "Send email to X" For each element, we define a name, anchor, text, and a list of parameters. It is worth noting the use of the operator "<<" called Merge Key Language-Independent Type (see Figure \ref{fourthsection}). This operator allows us to get the values defined in the previous section of our document and put them in the place indicated by the << sign. The same element defined early may thus be used in many places. The params element follows the element marked with <<. The order is essential here. Some of the items require parameters that will vary depending on where the item is used. Because we specified params after the << character, we overwrite the values taken from the element definition. In the next section, we define the process's starting point and endpoint and specify their identifiers. Thus, we can have one starting point and one ending point.

% In the next section, we define the process's starting point and endpoint and specify their identifiers (see table \ref{thirdsection}). Thus, we can have one starting point and one ending point.

\begin{figure}[hbt!]
  \centering
  \includegraphics[width=0.5\linewidth]{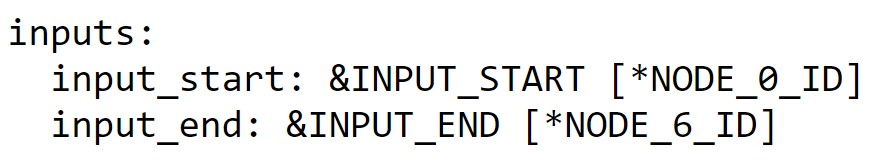}
  \caption{Section used to define starting and ending node}
  \label{thirdsection}
\end{figure}

Finally, the flow between individual elements is modeled (see Figure \ref{fourthsection}). As we can see, the first item is NODE\_0\_ID. The "references\_in" point to the previous items, and the "references\_out" properties point to the next elements in the process. In addition, we can pass the parameters needed to perform a given action in the action section. Here, with the help of <<, we refer to the previously defined blocks. We can also overwrite the parameter list. In this case, the file contains all the elements that should allow us to understand the described process. In addition, we are still providing the parameters necessary in a situation where we would like the language to continue to be machine-processable. In this way, in the future, we still have the possibility of creating an engine that would reproduce the process written in this language within a specific environment or a tool that would help us visualize the process.

\section{Conclusions}

Our proposed eXtensible Robotic Language (XRL), based on YAML, is a step towards creating a robotic process automation language that would be universal, human-readable, concise, and computer-processable. Its aim is also to standardize and unify the output code of different GUI-based RPA solutions, to prevent them from monopolizing the market. XRL could lower entry barriers for people with little programming experience, helping them participate in the automation of repetitive tasks using software robots. The proposed solution could facilitate the easy transfer of recorded processes between various tools and allow possible integration with tools such as ChatGPT for the automation of record keeping. This would allow people who were previously tied up in repetitive dead-end jobs to engage with more creative and motivating tasks. It could also help combat job loss that is due to automation, as the very people whose jobs are being automated could engage with automation tools more efficiently and help maintain them.

\bibliographystyle{splncs04}
\bibliography{bibliography}

\begin{thebibliography}{1}
\providecommand{\url}[1]{\texttt{#1}}
\providecommand{\urlprefix}{URL }
\providecommand{\doi}[1]{https://doi.org/#1}

\bibitem{2021YamlOfficialPage}
The {Official} {YAML} {Web} {Site} (2021), \url{https://yaml.org/}

\bibitem{chasins2021pl}
Chasins, S.E., Glassman, E.L., Sunshine, J.: Pl and hci: better together. Communications of the ACM  \textbf{64}(8),  98--106 (2021)

\bibitem{kopec2018hybrid}
Kope{\'c}, W., Skibi{\'n}ski, M., Biele, C., Skorupska, K., Tkaczyk, D., Jaskulska, A., Abramczuk, K., Gago, P., Marasek, K.: Hybrid approach to automation, rpa and machine learning: a method for the human-centered design of software robots. arXiv preprint arXiv:1811.02213  (2018)

\bibitem{codereadability}
Oliveira, D., Bruno, R., Madeiral, F., Castor, F.: Evaluating code readability and legibility: An examination of human-centric studies. In: 2020 IEEE International Conference on Software Maintenance and Evolution (ICSME). pp. 348--359 (2020). \doi{10.1109/ICSME46990.2020.00041}

\bibitem{codereadabilitytesting2016}
Sedano, T.: Code readability testing, an empirical study (04 2016). \doi{10.1109/CSEET.2016.36}

\bibitem{criteriaforevaluation2016}
Sheikh, G., Islam, N.: A qualitative study of major programming languages: teaching programming languages to computer science students. International Journal of Information and Communication Technology  (01 2016)

\bibitem{codereadability2020}
Tariq, M.U., Bashir, M., Babar, M., Sohail, A.: Code readability management of high-level programming languages: A comparative study. International Journal of Advanced Computer Science and Applications  \textbf{11},  595--602 (03 2020). \doi{10.14569/IJACSA.2020.0110375}

\bibitem{willcocks_paper_nodate}
Willcocks, P.L.: Paper 15/05 {The} {IT} {Function} and {Robotic} {Process} {Automation} p.~39 (Oct 2015)

\end{thebibliography}

\end{document}